\documentclass[conference]{IEEEtran}
\IEEEoverridecommandlockouts

\usepackage[ruled,vlined,linesnumbered]{algorithm2e}
\SetKwInput{KwData}{Input}
\SetKwInput{KwResult}{Output}

\usepackage{subfigure}
\usepackage{enumitem}
\usepackage{hyperref}
\usepackage{caption}
\usepackage{cite}
\usepackage{amsmath,amssymb,amsfonts}
\usepackage{algorithmic}
\usepackage{graphicx}
\usepackage{textcomp}
\usepackage{xcolor}
\usepackage[a4paper, total={184mm,239mm}]{geometry}
\def\BibTeX{{\rm B\kern-.05em{\sc i\kern-.025em b}\kern-.08em
    T\kern-.1667em\lower.7ex\hbox{E}\kern-.125emX}}

\usepackage{etoolbox}

\begin{document}

\title{NetCut: Real-Time DNN Inference Using Layer Removal}

\author{\IEEEauthorblockN{Mehrshad Zandigohar, Deniz Erdo\u{g}mu\c{s}, Gunar Schirner}
\IEEEauthorblockA{\textit{Department of Electrical and Computer Engineering, Northeastern University, USA } \\
\{zandi, erdogmus, schirner\}@ece.neu.edu \vspace{-8pt}}
}

\maketitle
\global\csname @topnum\endcsname 0
\global\csname @botnum\endcsname 0

\begin{abstract}
Deep Learning plays a significant role in assisting humans in many aspects of their lives. As these networks tend to get deeper over time, they extract more features to increase accuracy at the cost of additional inference latency. This accuracy-performance trade-off makes it more challenging for Embedded Systems, as resource-constrained processors with strict deadlines, to deploy them efficiently. This can lead to selection of networks that can prematurely meet a specified deadline with excess slack time that could have potentially contributed to increased accuracy. 

In this work, we propose: (i) the concept of layer removal as a means of constructing TRimmed Networks (TRNs) that are based on removing problem-specific features of a pretrained network used in transfer learning, and 
(ii) NetCut, a methodology based on an empirical or an analytical latency estimator, which only proposes and retrains TRNs that can meet the application's deadline, hence reducing the exploration time significantly. 

We demonstrate that TRNs can expand the Pareto frontier that trades off latency and accuracy to provide networks that can meet arbitrary deadlines with potential accuracy improvement over off-the-shelf networks. Our experimental results show that such utilization of TRNs, while transferring to a simpler dataset, in combination with NetCut, can lead to the proposal of networks that can achieve relative accuracy improvement of up to 10.43\% among existing off-the-shelf neural architectures while meeting a specific deadline, and 27x speedup in exploration time.

\end{abstract}


\section{Introduction}
Deep Learning has shown promising results in assisting humans with a variety of tasks in the domains of classifying and detecting objects, processing natural languages and translations, speech recognition, and many more \cite{deeplearning, speech, alexnet}. Recently, Deep Neural Networks (DNN) have been applied to several critical domains including medical diagnosis and monitoring, autonomous driving, and assistive robotics \cite{medical,auto,vivian}, which generally use a pretrained network for the targeted task.

\begin{figure}[t]
  \centering
  \includegraphics[width=1\linewidth]{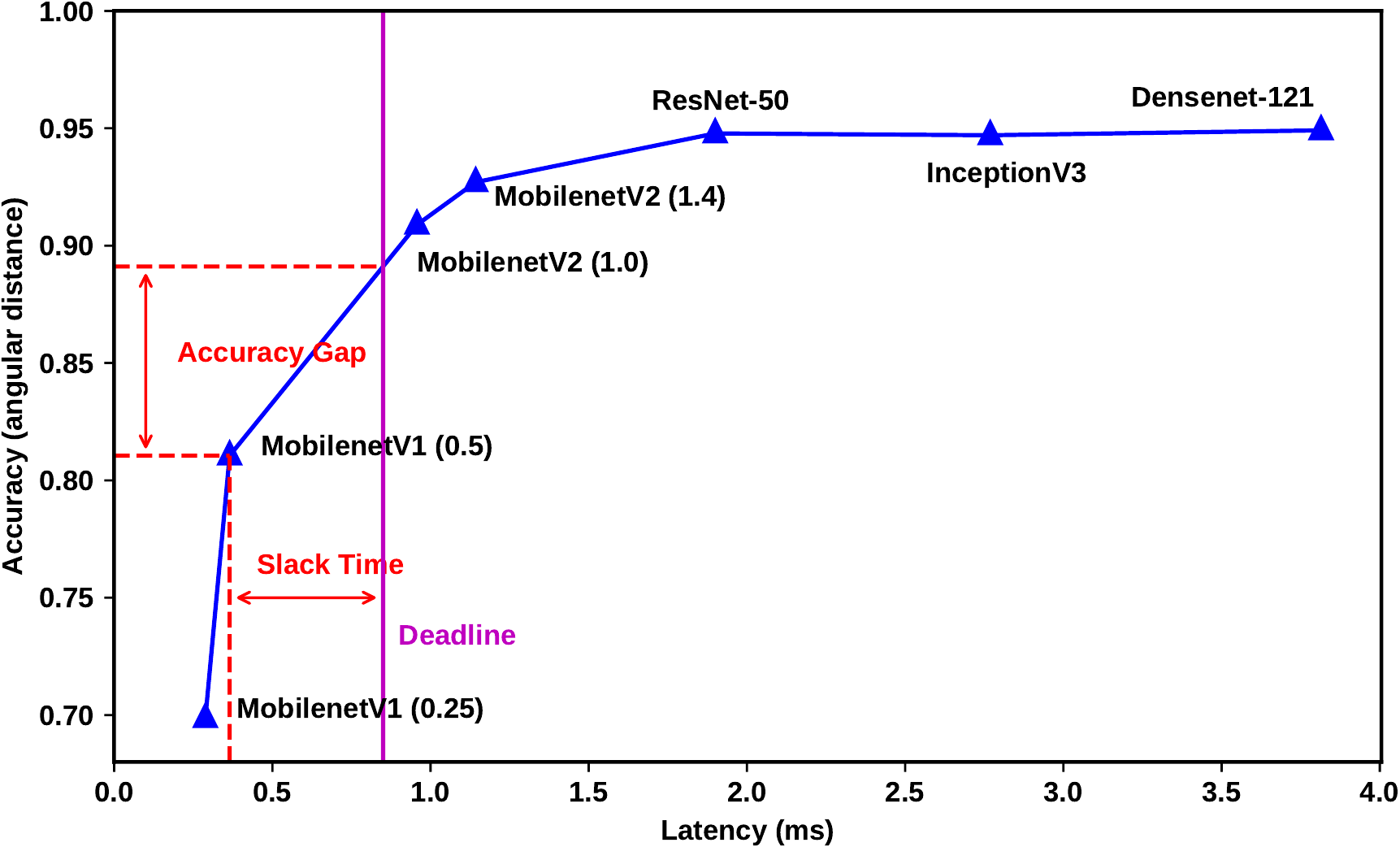}
  \caption{The trade-off between latency and accuracy.}
  \label{fig:pareto0}
\end{figure}

With the rise of DNNs beginning from AlextNet \cite{alexnet} with 8 layers, later VGG \cite{vgg} with 19 layers, to the more recent and advanced networks like ResNet \cite{resnet50} or DenseNet \cite{densenet} with over 100 layers; DNNs are generally getting deeper and extracting more features while adding to their inference latency \cite{vivian}. This increasing growth of network sizes has made it challenging for embedded devices to employ the state-of-the-art neural architectures because of their tendency to violate the deadline necessary to perform a task correctly. As a result, deploying them demands careful attention to the accuracy and latency trade-off. Networks with a longer latency generally show increased accuracy, or otherwise are dominated by other networks. This results in a Pareto frontier defining the trade-off as depicted in \autoref{fig:pareto0}. When choosing a network, the goal is to select the most accurate network that meets the specified deadline. While prematurely meeting deadlines is desirable from a real-time execution perspective, this creates an accuracy gap because the slack time does not contribute to increasing the accuracy. Hence, the additional slack time should be used to increase network accuracy (\autoref{fig:pareto0}).

To this end, we propose layer removal to construct TRimmed Networks (TRNs), that expands the Pareto frontier and increase diversity among candidate networks for a given task. In addition, we propose NetCut, a methodology which utilizes layer removal but predicts the latency of TRNs, with the aim of selecting only the networks that can meet the desired deadline. This results in much fewer TRNs to train thus speeding up exploration time. Layer removal is based on the intuition that most of the features at the beginning of a network extract high-level features (like edges and colors) which is common among many image classification tasks. In contrast to the initial layers, last layer features are more problem-specific \cite{howtransferable} and are not necessarily useful enough when transferring knowledge from a pretrained network to a new task, especially if the required task is simpler than that of the original task. This makes the last layers more suitable for removal. 

To demonstrate the efficacy of NetCut, and with the aim of assisting amputees with their daily life activities, the experiments in this paper are conducted on a robotic prosthetic hand. The robotic hand application is used as the example application for NetCut. We use 7 efficient pretrained networks from ImageNet for the task of estimating the probability of grasp types executed on an embedded GPU. Using layer removal, we show that while meeting the robotic hand application deadline, we can find appropriate TRNs with up to 10.43\% accuracy improvement over off the shelf networks while increasing exploration time by 27x.

The main contributions of this work are:
\begin{itemize}[noitemsep,topsep=0pt]
    \item \textbf{Extensive study on layer removal:} the effects of removing layers on accuracy and performance of several networks and proposing blockwise removal as a suitable heuristic to narrow the removal search space.
    \item \textbf{NetCut:} A fast layer removal exploration methodology based on network latency estimation models, both analytical and profiler based, to find the most accurate TRN that meets an arbitrary deadline.
\end{itemize}


\section{Related Work}
There has been extensive research on co-design of DNN models and hardware \cite{vivian}, including methods which utilize quantization, sparsity and pruning \cite{QNN, struct_pruning, mobilenet}. However, to the best of our knowledge, this work is the first study on layer removal, and unlike our work, these methods focus on accelerating individual networks and do not propose a methodology to select efficient networks from a wide range of neural architectures. As a result, our work can augment the aforementioned studies by pruning the search space and providing an efficient network from a diverse set of networks which can later be used by the current literature.

Similar to our work, there has been several studies based on the principle of transferability of features in DNNs \cite{BranchyNet, edgent}. BranchyNet \cite{BranchyNet} introduces the concept of early exiting to use more general layers at runtime based on the confidence of each exit point on a single architecture \cite{BranchyNet}. Ours on the other hand, expands the proposed networks based on multiple networks. 

Edgent \cite{edgent}, another related methodology based on latency prediction, tries to predict the latency of networks based on linear regression models for different layer types to partition network computation demand between server and the edge. Our proposed method for latency prediction, on the other hand, uses a coarser-grained analytical model to enable employing other optimizations i.e. layer fusion which is impossible using models based on layer types.

NetAdapt \cite{NetAdapt}, also aims to adapt a pretrained network to meet an arbitrary deadline through pruning. However, it focuses on a single individual network and requires retraining in each iteration of its algorithm to evaluate network proposals. In result, it suffers from a long exploration time making it impractical to be applied to a diverse set of networks. 

In conclusion, NetCut as an orthogonal approach, can augment the existing methods by exploring a diverse set of networks and providing the most efficient in a reasonable time.

\section{Application Example}
Embedded systems control many aspects of human life today throughout the world. They have been extensively used in numerous industrial, automotive, medical, and many more applications. Among them, assistive robots including robotic prosthetic hands, try to compensate the lost ability of amputee's upper limbic system through processing different sensory data. The robotic hand as part of a control loop, is expected to perform accurately and in time. To provide background and demonstrate the real-time challenges of the robotic hand, we will outline the robot's system overview, training setup and performance in the following subsections. The robotic prosthetic hand used in this work is adapted from \cite{cyphy19}.

\begin{figure}[t]
  \centering
  \includegraphics[width=1\linewidth]{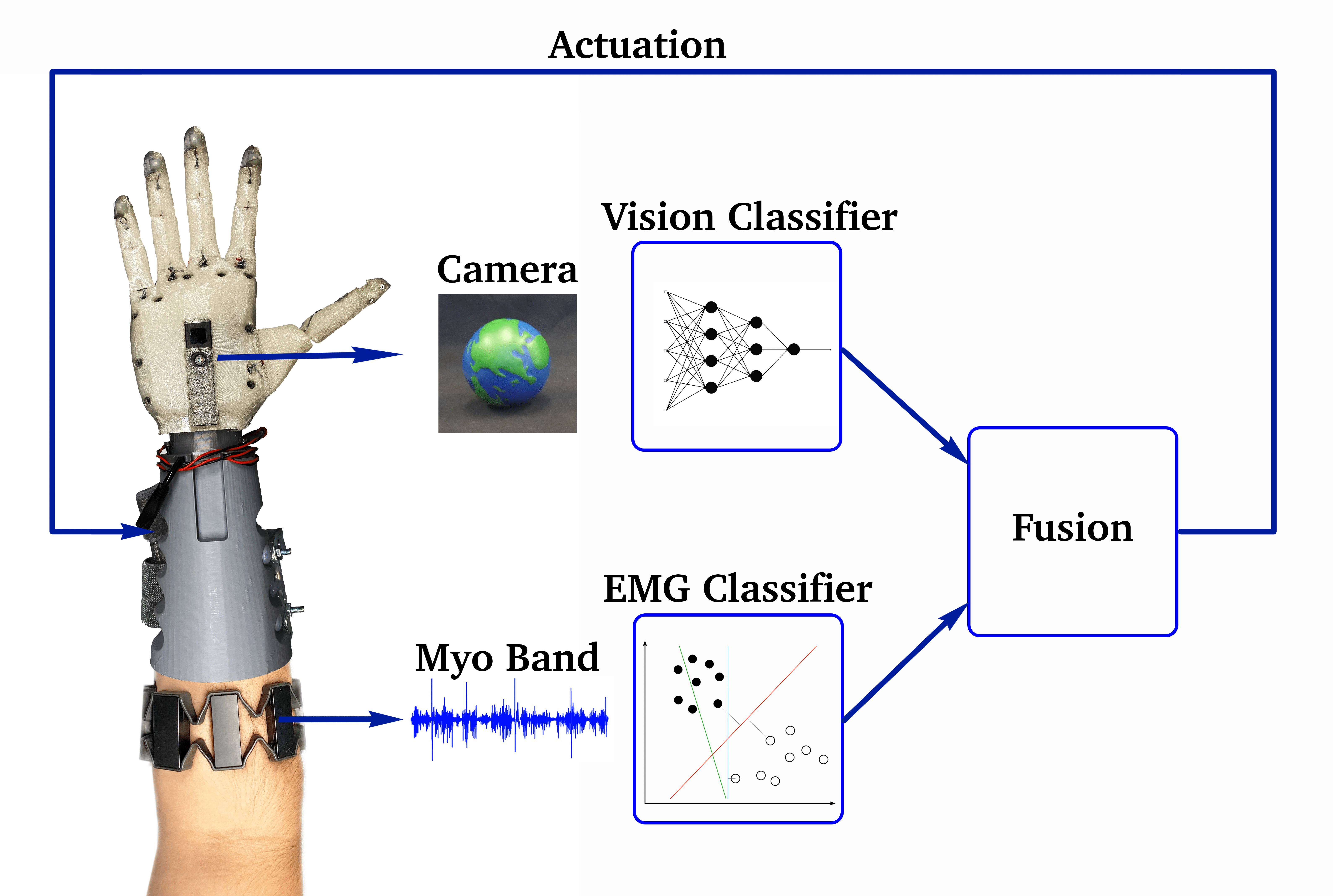}
  \caption{System overview of the robotic prosthetic hand.}
  \label{fig:overview}
\end{figure}

\subsection{Robotic Hand}
The control loop of the robotic hand is illustrated in \autoref{fig:overview}. The robot consists of an EMG classifier to infer human intent using the amputee's remaining limbic signals with a Myo band, and a convolutional neural network using the images from a palm camera. The use of a visual classifier in the control loop is crucial as relying solely on EMG signals lacks robustness and yields poor results. In order to fuse the predictions from the aforementioned classifiers, each one produces a probability distribution of probable grasp types, in contrast to the one-hot encoding, which results in a final decision. The accuracy is therefore measured using angular similarity. Using the decision from fusion, actuation signals are sent to the robotic hand to form the correct hand gesture.

During the course of reaching for an object, the final decision should have been made prior to the contact with the intended object, giving enough time to the actuation unit to perform the intended grasp type. During this limited time, fusion adds reliability by using several predictions prior to the final decision which further tightens the deadline. From our experiments, we observed that given all the system constraints and design parameters, the visual classifier needs to predict within 0.9 ms of receiving a frame and preprocessing it prior to writing back to the main memory. 

\subsection{Visual Classifier}
Because of the tight deadline which the visual classifier needs to meet, it is essential to choose the most accurate model which meets the timing requirement. This subsection outlines the details for selecting, retraining and optimizing efficient deep neural networks for the visual classifier.

\subsubsection{Select Efficient Pretrained Networks}
\label{sec:eff_selection}
To select efficient neural networks which can both be accurate and meet the specific deadline from the numerous pretrained networks in the field is challenging. Authors in \cite{cyphy19} try to select efficient sources of transfer by analyzing the top-5 accuracy and FLOPs from 23 off-the-shelf ImageNet pretrained networks. They eliminate the models which are dominated by the models on the Pareto frontier to reduce the search space. As suggested, our work also uses MobileNetV1 (0.25, 0.5), MobileNetV2 (1.0, 1.4), and InceptionV3, in addition to ResNet50, and DenseNet-121 for training \cite{mobilenet,inceptionV3,resnet50,densenet}.

\subsubsection{Dataset}
The HANDS dataset \cite{han2020hands} is a collection of images from graspable objects used in daily life including office supplies, utensils and complex-shaped objects like toys, from the hand camera perspective and different orientations. The labels are probabilistic as opposed to the common one-hot encoding because of the feasibility of grasping objects in multiple ways with different preferences. The labels consist of 5 grasp types i.e. Open Palm, Medium Wrap, Power Sphere, Parallel Extension, and Palmar Pinch.

\subsubsection{Fine-tuning and evaluation}
To construct the models for transfer learning, the top Fully Connected (FC) layers a.k.a. classification layers are removed and replaced with 1 Global Average Pooling to reduce the dimensionality of output shape, 2 FC/ReLU layers, and 1 FC/Softmax layers for mapping features to custom classes (bottom-right part of \autoref{fig:CNN_layer}). The training starts with the initial learning rate of $10^{-3}$ while all the features are frozen and then continued for another 50 epochs with a lower learning rate of $10^{-4}$ while all the layers are updated. Due to the results being a probability distribution, accuracy definitions as used in one-hot encoded classifiers are not applicable. Instead, angular distance has been shown to provide a more intuitive and accurate metric \cite{cyphy19}. 

\subsubsection{Deployment Optimizations}
To optimize the trained networks for inference, we use post-training quantization \cite{quantization} of weights and activations to exploit the fast integer arithmetic operations. The weights are quantized per-feature offline and the activations are quantized on the fly per-tensor. To collect the statistics of the activations, 10\% of the training set is randomly selected as the calibration set. Thereafter, calibration images are fed to the network and scaling factors which minimize the information loss are selected. Also, layer fusion and MobileNets with different multipliers have been used.

\subsection{The Gap in Off-The-Shelf Networks}
\label{sec:gap}

\autoref{fig:pareto0} shows the latency and accuracy of the selected architectures on the robotic hand. To meet the 0.9 ms deadline imposed on the visual classifier, MobileNetV1 (0.5) can achieve an accuracy of 0.81 with an inference delay of 0.36 ms. As depicted with a vertical line in \autoref{fig:pareto0}, there is a gap in accuracy between the deadline and the selected network. Although there is a slack time with possible improvement in the accuracy, simply relying on the off-the-shelf networks prevents this opportunity. In the next section, we will provide evidence that layer removal can increase candidate networks that will exploit the slack time to provide more accuracy while meeting the specified deadline.

\section{Layer Removal}
\label{sec:layer_removal}

In practice, most of the Convolutional Neural Networks are not trained from scratch due to the scarcity of having a large enough dataset to begin with. To this end, transfer learning has proved to be highly effective when a CNN is pretrained on a very large dataset, e.g. ImageNet, with millions of images and numerous categories, and fine-tuned to the new dataset \cite{decaf}. 

The features learned by the network, vary significantly in their characteristics with respect to their topological order \cite{howtransferable}. The earlier layers contain features similar to Gabor filters and color blobs which try to distinguish low-level features like edges and colors which are generally found in CNNs of many computer vision tasks. However, later layers become progressively more specific to the task in the original dataset. Moreover, since the original dataset is often much more complex than the target task, the sophisticated features extracted from later layers might not be useful for simpler tasks. 

Based on these observations, layer removal tries to construct low latency trimmed networks by removing problem-specific features and as a result expanding the Pareto frontier. This section will study the effects of layer removal on accuracy, performance, and the resulting Pareto frontier.

\subsection{Narrowing the Exploration Space}
\begin{figure}[t]
  \centering
  \includegraphics[width=0.9\linewidth]{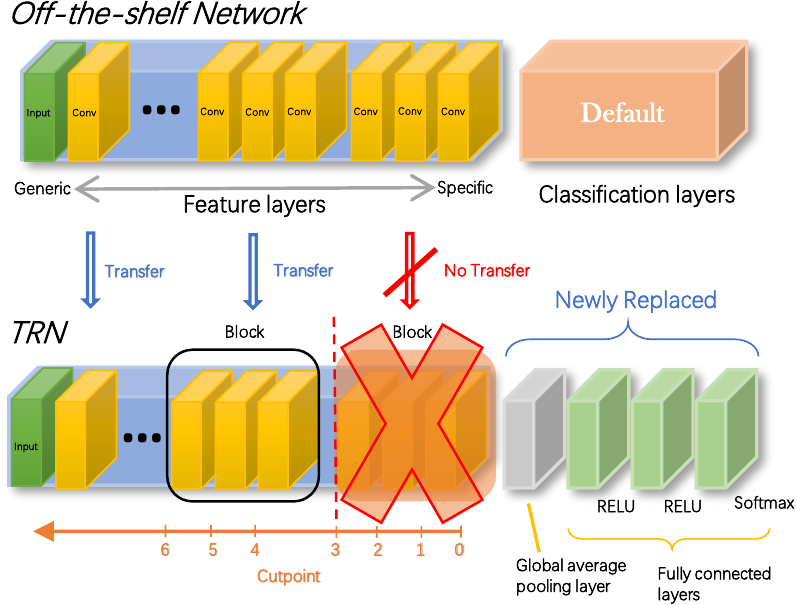}
  \caption{Constructing TRN using blockwise layer removal.}
  \label{fig:CNN_layer}
  \vspace{6pt}
\end{figure}
\begin{figure}[t]
  \centering
  \includegraphics[width=0.9\linewidth]{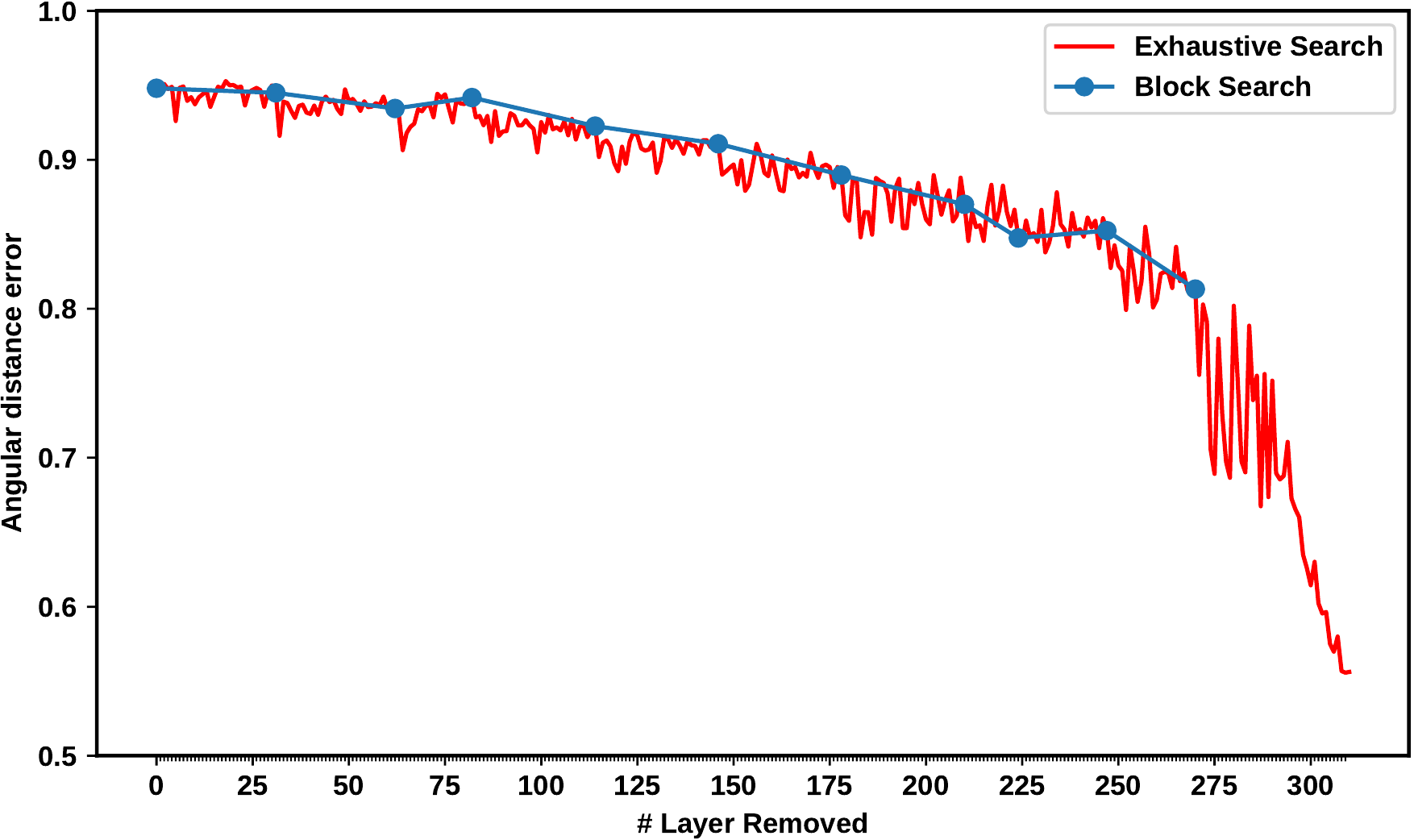}
  \caption{\small Blockwise layer removal compared with iteratively removing each layer for InceptionV3.}
  \label{fig:blockwise}
  \vspace{-4pt}
\end{figure}

\begin{figure*}[t]
  \centering
  \begin{minipage}[t]{0.32\textwidth}
    \includegraphics[width=0.95\textwidth]{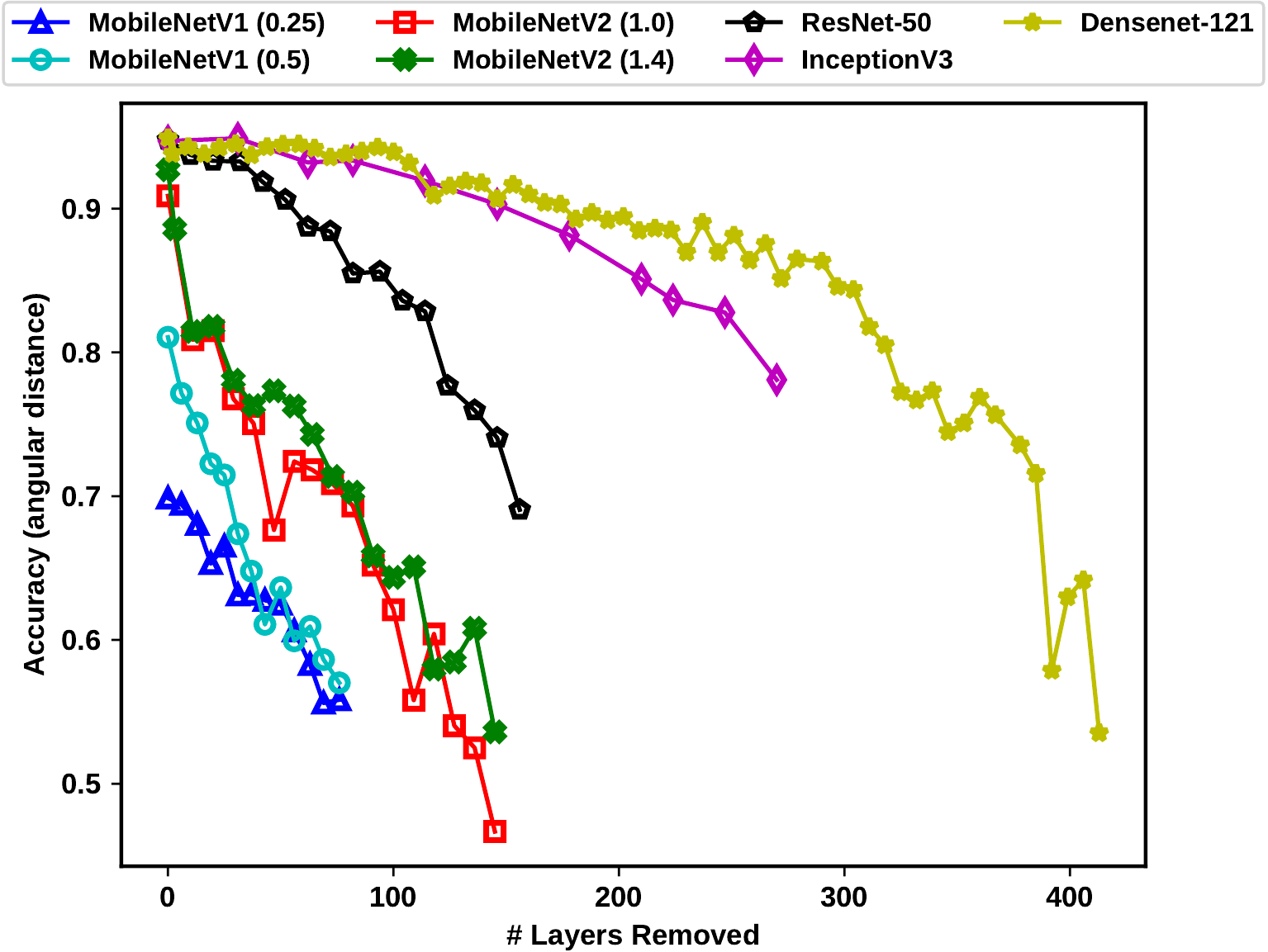}
        \vspace{-2pt}
    \caption{\small The effects of layer removal on accuracy of different architectures.}
    \label{fig:acc8}
  \end{minipage}
  \begin{minipage}[t]{0.32\textwidth}
    \includegraphics[width=0.95\linewidth]{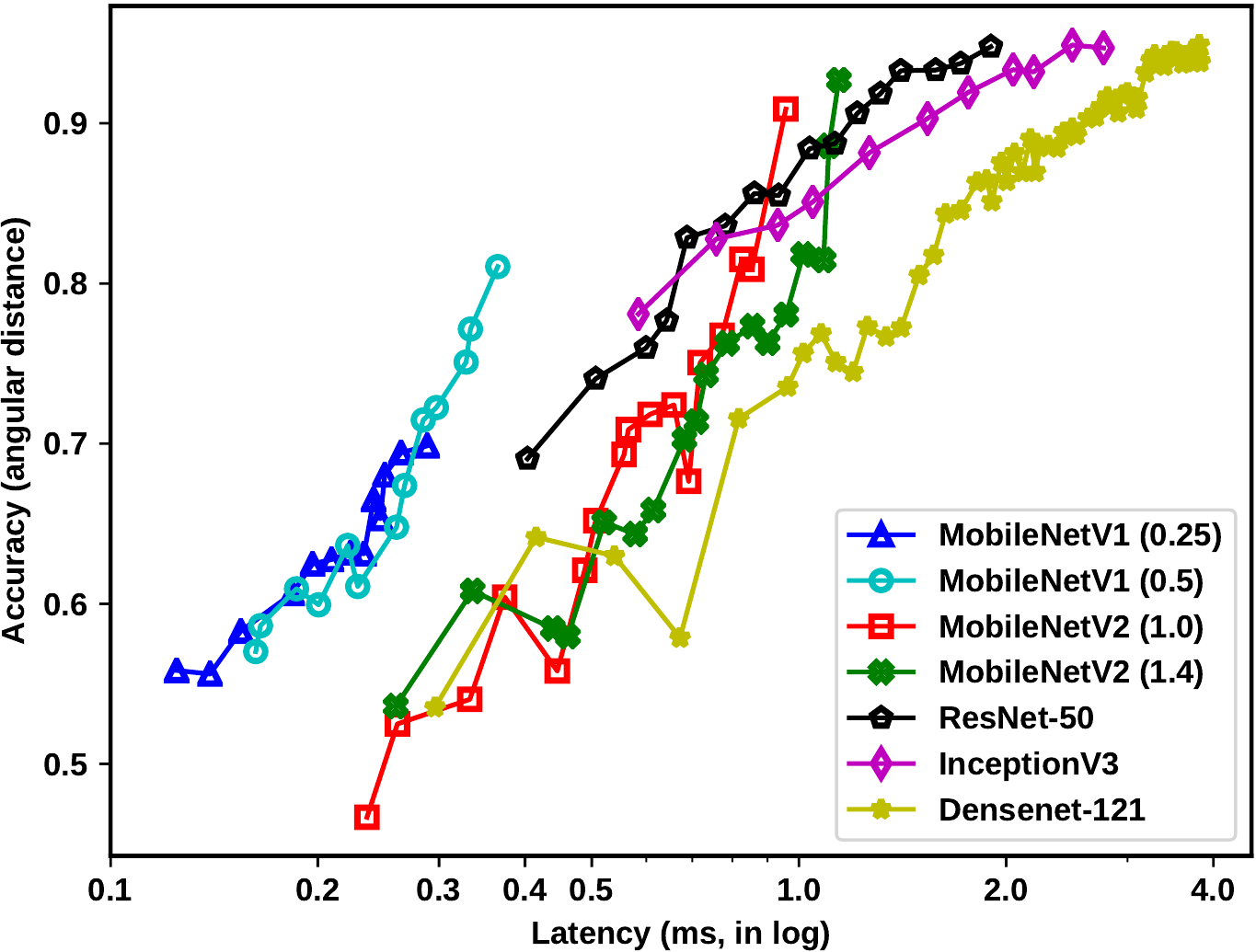}
        \vspace{-2pt}
    \caption{\small Accuracy-Performance trade-off of TRNs using blockwise layer removal.}
    \label{fig:new_trade}
  \end{minipage}%
  \hfill
  \begin{minipage}[t]{0.32\textwidth}
    \includegraphics[width=0.95\linewidth]{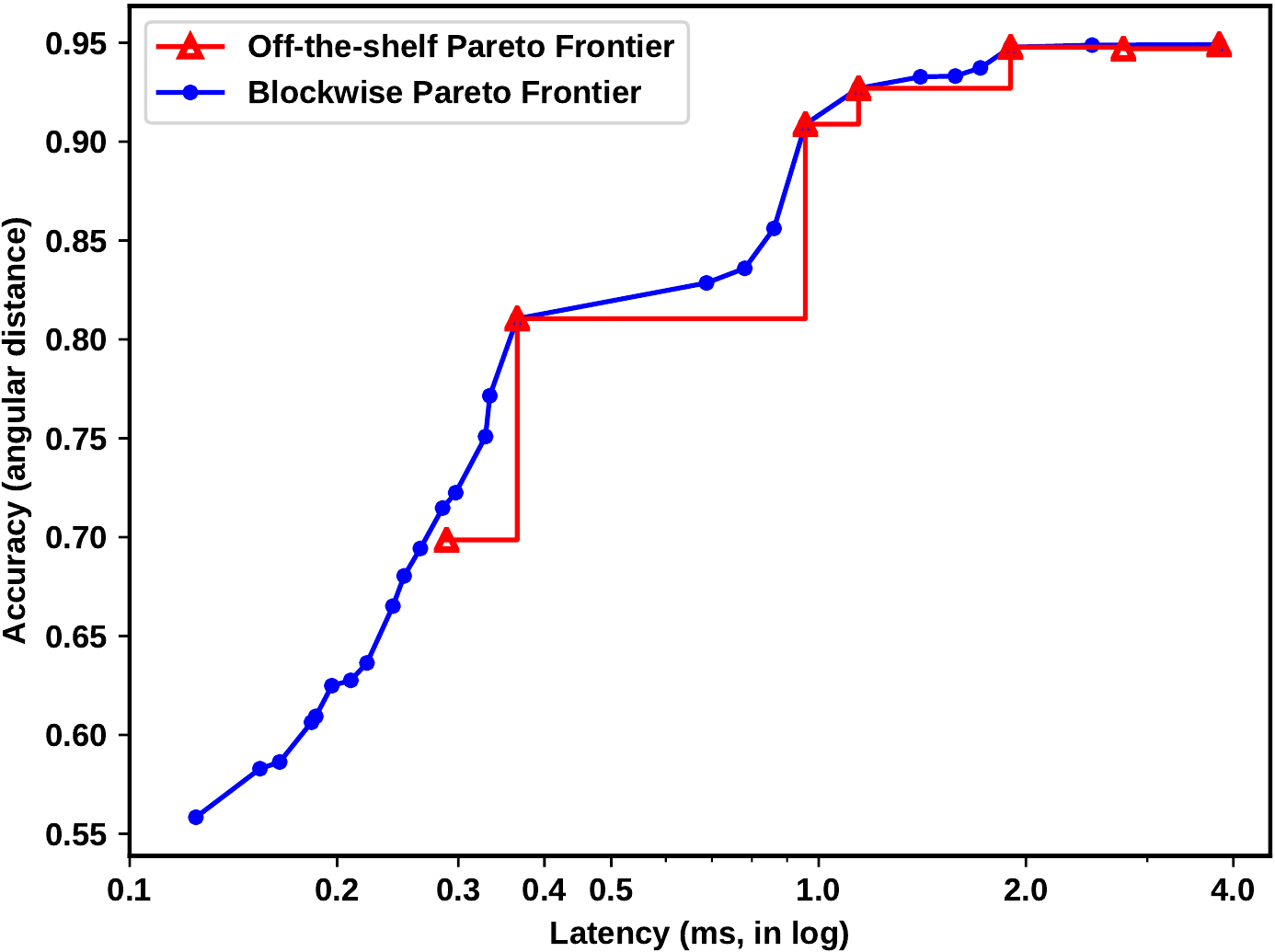}
        \vspace{-2pt}
    \caption{\small Estimation Accuracy}
    \label{fig:improv}
  \end{minipage}
  
\end{figure*}

Convolutional networks are often constructed from several blocks which are repeating modules in the network structure that encapsulate filters with different sizes. In \cite{inceptionV3}, the authors illustrate several examples of blocks, a.k.a. Inception modules throughout their design. Also, we can find the same concept in \cite{resnet50}, as the authors describe residual blocks. With the use of blocks, the network can learn variations in the same category by capturing spatial information at different granularity levels.  

As iteratively removing layers from networks with hundreds of layers can significantly increase the number of TRNs, to further study the effects of layer removal, we will use the concept of blockwise layer removal to narrow our exploration space (bottom-left of \autoref{fig:CNN_layer}). This will improve the exploration time, and provides a suitable heuristic for exploration. \autoref{fig:blockwise} compares blockwise layer removal with iterative layer removal (exhaustive search) for InceptionV3. During our experiments, we observed that removing the whole block instead of keeping some of the layers inside that block often has none or negligible accuracy loss (\textless0.03). Therefore, we will use block-level granularity for layer removal.

\subsection{Accuracy-Performance Trade-off}
\subsubsection{Effects on Accuracy}
To measure how layer removal affects the accuracy of different networks, we start with networks from \autoref{sec:eff_selection}, construct TRNs using blockwise layer removal and finally train and evaluate them resulting in 148 networks in total. As \autoref{fig:acc8} shows, some networks are more sensitive to layer removal than others. We can observe that DenseNet and Inception exhibit a low loss in accuracy passing more than 100 layers removed and start to smoothly drop afterwards. This suggests that the deeper networks might have more problem-specific features. On the contrary, the accuracy in MobileNets are drastically dropped with the slightest layer removal. Another observation is that while ResNet and MobileNetV2 have nearly the same number of layers, MobileNetV2 is more adversely affected by layer removal. This indicates that the features in the neural architectures other than MobileNets (regardless of their hyper-parameters including depth and width multipliers or network depth) are more generalized and better transferred.

\subsubsection{Effects on Performance}
The embedded platform used in our work is NVIDIA's Jetson Xavier. To have consistent performance results, we first warm-up the GPU by running 200 inferences and then report the inference latency as the average over another 800 runs. For each inference, the compute time starts right after the inputs are transferred until they are ready to be transferred back to the host. As expected, the results demonstrate that inference latency decreases almost linearly w.r.t. to the number of layers removed. Since the results were intuitive, we have excluded the figures for this part.

\subsection{The New Pareto Frontier}
Combining accuracy and performance measurements of the TRNs yields a new trade-off depicted in \autoref{fig:new_trade}. We can observe that ResNet, can provide fairly accurate TRNs which fill the gap prior to MobileNetV2 (1.4). On the other side of the figure, when analyzing smaller networks, removing layers from MobileNetV1 (0.5) can also greatly expand the Pareto frontier in order to use the extra slack time to recover accuracy. Another interesting observation is that layer removal on MobileNetV1 (0.5) can dominate the off-the-shelf MobileNetV1 (0.25).

Using the accuracy-performance trade-off, the new Pareto frontier is extracted and depicted in \autoref{fig:improv}. In most cases, layer removal has provided TRNs that can contribute to increased accuracy using the allocated slack time. As an example, removing 1 block from MobileNetV1 (0.5) will yield a TRNs with a 10.43\% relative increase in accuracy. Our experiments show that on average, the relative accuracy improvement for all TRNs is 5.0\% for our application. Another observation is that layer removal can also expand the Pareto to the lower extreme of the trade-off.

Witnessing the ability of layer removal in utilizing the slack time to increase network diversity with TRNs, and hence recovering the accuracy, we will introduce our proposed methodology, NetCut, in the following section. Based on deadline information, NetCut only selects TRNs which can meet the deadline, leading to significantly lower exploration time.

\section{NetCut: Deadline-Aware Exploration}
As discussed in \autoref{sec:layer_removal}, layer removal is able to utilize slack time and construct TRNs with potentially higher accuracy. However, blockwise removal requires days if not weeks of retraining the TRNs and executing them on the targeted device to evaluate their performance. To have a practical and more efficient exploration, we propose NetCut, a methodology based on layer removal which significantly reduces the exploration time. NetCut is based on the observation that generally, among TRNs from the same initial network, the most accurate ones meeting the deadline are the ones close to it as they utilize more parameters an therefore yield more accurate networks. Hence, based on a latency estimator, NetCut only trains and evaluates the TRN with the latency closest to the targeted deadline. This means that for our application with 7 initial networks, only 7 TRNs are constructed, retrained, and evaluated, which compared to 148 candidates from blockwise layer removal, is a 95\% reduction in number of networks. The following sections present the details.

\subsection{Algorithm Overview}
The algorithm (\autoref{alg:netcut}) starts with one of the $N$ trained off-the-shelf networks ($Net_i$), and iteratively increments the cutpoint to remove layers from the top of the network and construct a TRN($TRN_i$). Afterward, the latency of the TRN will be estimated to ensure meeting the desired deadline ($Est\_Latency$). The algorithm continues until the very first real-time TRN is proposed; thereafter, the algorithm repeats the same actions for the remaining off-the-shelf networks until there is one real-time TRN for each neural architecture ($N$ TRNs in total). Then, the TRNs will be retrained ($Trained\_TRN$) to evaluate accuracy, and the algorithm terminates after picking the one with the highest accuracy ($\overline{Net}$).

In order for NetCut to be effective, it is essential to perform accurate latency estimation. The next section describes two approaches for predicting inference latency of the newly constructed models. 

\setlength{\textfloatsep}{10pt}
\begin{algorithm}[t]
\small
\SetAlgoLined
\KwData{$N$ Trained networks: $Net_i$, $N$ Inference latency measurements: $Latency_i$, $N$ Accuracy measurements: $Accuracy_i$, Deadline}
\KwResult{Efficient network meeting the deadline: $\overline{Net}$}
 
 \For{$i\gets1$ \KwTo $N$}{
$TRN_i = Net_i$\;
$Est\_Latency = Latency_i$\;
 $Cutpoint=1$ \;
 \While{$Est\_Latency > Deadline$}{
  $TRN_i=RemoveLayers(Net_i, Cutpoint)$ \;
  $Est\_Latency = EstimateLatency(TRN_i, Latency_0)$\;
  $Cutpoint \leftarrow Cutpoint + 1$ \;
 }
 $Trained\_TRN_i = Retrain(TRN_i)$ \;
}
$\overline{Net} = PickHighestAccuracy(Trained\_Net)$ \;
return $\overline{Net}$ \;
\caption{NetCut algorithm}
\label{alg:netcut}
\end{algorithm}
\subsection{Latency Estimation}
The latency estimator uses layer statistics from each initial network to predict the TRN's inference latency. For this, we propose two approaches: (1) a profiler-based estimation measuring each layer's latency on the device, and (2) an analytical model that uses device-agnostic high-level features to train a regression model which can estimate the TRNs's performance.

\subsubsection{Profiler-Based Estimation}
\label{sec:profiler}
This approach aims to estimate the latency of a TRN by running and profiling the layers of the original network on the device. This is implemented using CUDA events which record the latency of executing each layer into a table. Hence, the number of tables generated is equal to the number of unmodified networks. To estimate a TRN's inference latency, this method looks into its appropriate table and subtracts the original architecture's latency by the ratio of the latency of the layers which do not exist anymore over the summation of all layers' latencies. The reason we decided to use a ratio is that we realized that in all cases, the summation of layers is slightly more than the actual measured inference delay. The reason can be explained by the overhead imposed by using CUDA events to measure each layer. The mathematical formulation of the profiler-based estimation can be denoted as: 
\begin{equation}
Latency(TRN_n) = Latency(Net_0) \cdot \\ 
( 1 - \frac{\sum_{i=0}^{n} {Latency(Layer_i)}}{\sum_{i=0}^{N} {Latency(Layer_i)}} )
\end{equation}

Where $Net_0$ is the initial network, $N$ is the total number of layers excluding classification layers, and $n$ is the desired cutpoint. Therefore, $Latency(TRN_n)$ is the measured inference latency of a TRN with $n$ removed layers, and $Latency(Layer_i)$ the latency of the $i^{th}$ layer starting from the last classification layer. Given the 7 networks that this work studies, profiler-based estimation only needs to construct 7 tables to estimate the performance of any TRN.

\subsubsection{Analytical Model}
Another approach proposed to estimate the performance of a network is to use device-agnostic, high-level features, to form an analytical model. This is especially useful when profiling information is not easily available. During our experiments, we observe that for a given network, the original network's latency, the total number of: floating-point operations, parameters, layers, and filter sizes will yield an accurate enough model to estimate the inference latency. The analytical model is then formulated as a Support Vector Regression ($\epsilon-SVR$) using Radial Basis Function (RBF) kernel. The hyper-parameters $\gamma=10^{-1}$ denoting kernel coefficient and $C=10^6$ or the regularization parameter were tuned using 10-Fold cross-validation on the train set and the final model was tested on the 80\% remaining data. We found that grid search outperforms random search in tuning the hyper-parameters as the sample size was not huge.

\subsection{Results}

\begin{figure*}[ht]
  \centering
  \begin{minipage}[t]{0.33\textwidth}
    \includegraphics[width=0.95\textwidth]{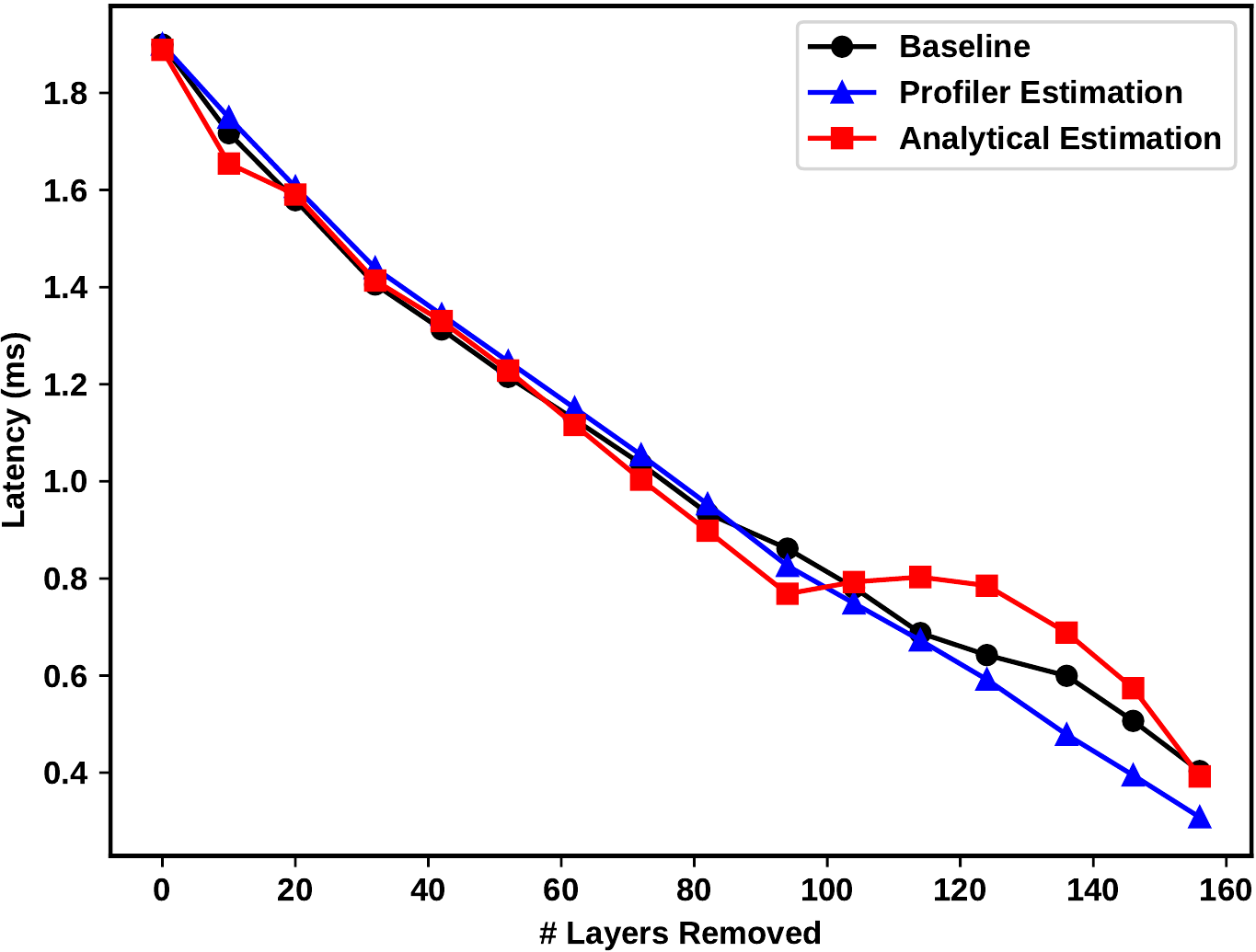}
    \caption{\small Estimations vs. Ground Truth (ResNet).}
    \label{fig:res_est}
  \end{minipage}%
  \hfill
  \begin{minipage}[t]{0.33\textwidth}
    \includegraphics[width=0.95\textwidth]{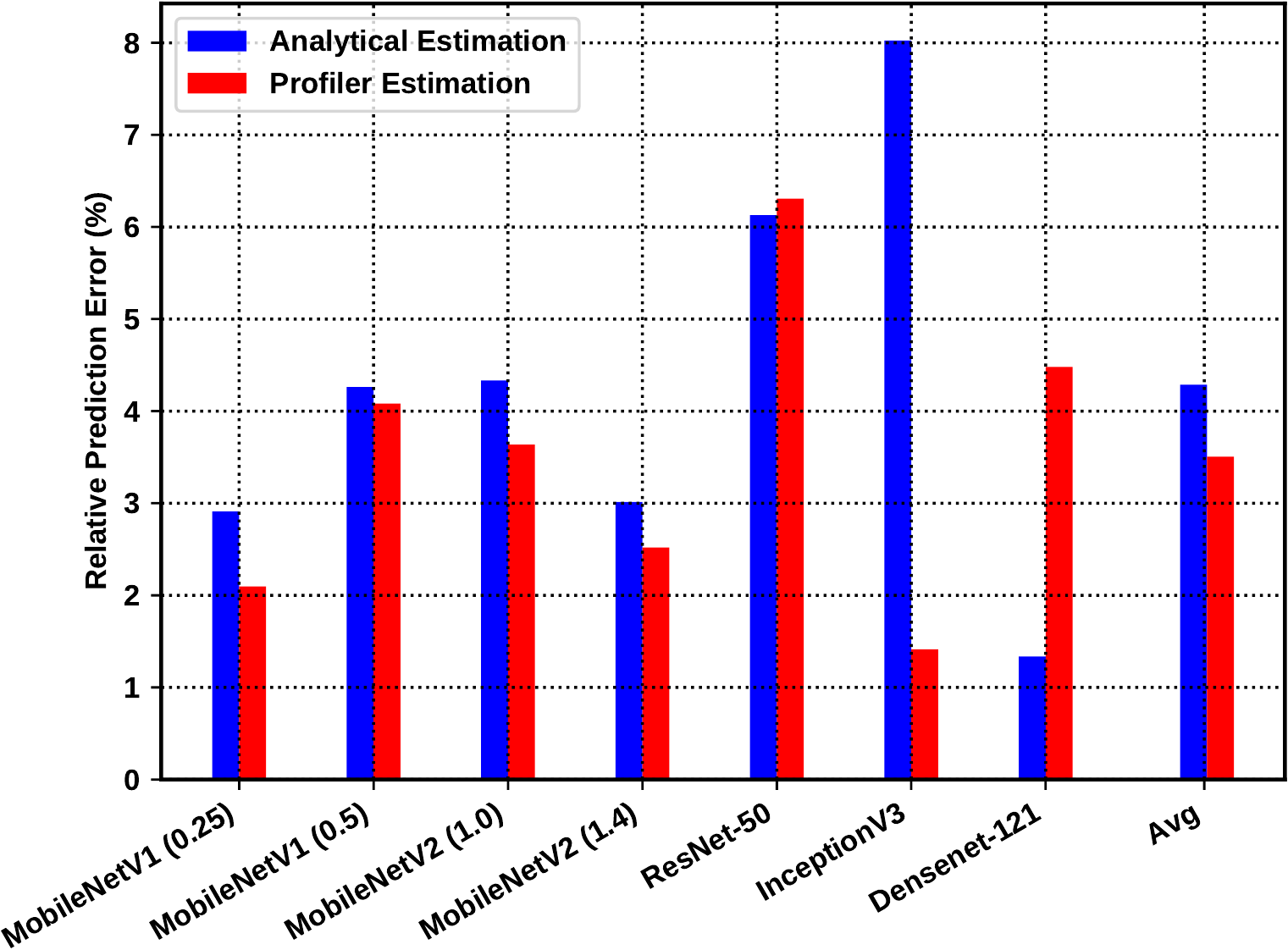}
    \caption{\small Estimation Accuracy}
    \label{fig:est_erros}
  \end{minipage}
  \begin{minipage}[t]{0.33\textwidth}
    \includegraphics[width=0.95\textwidth]{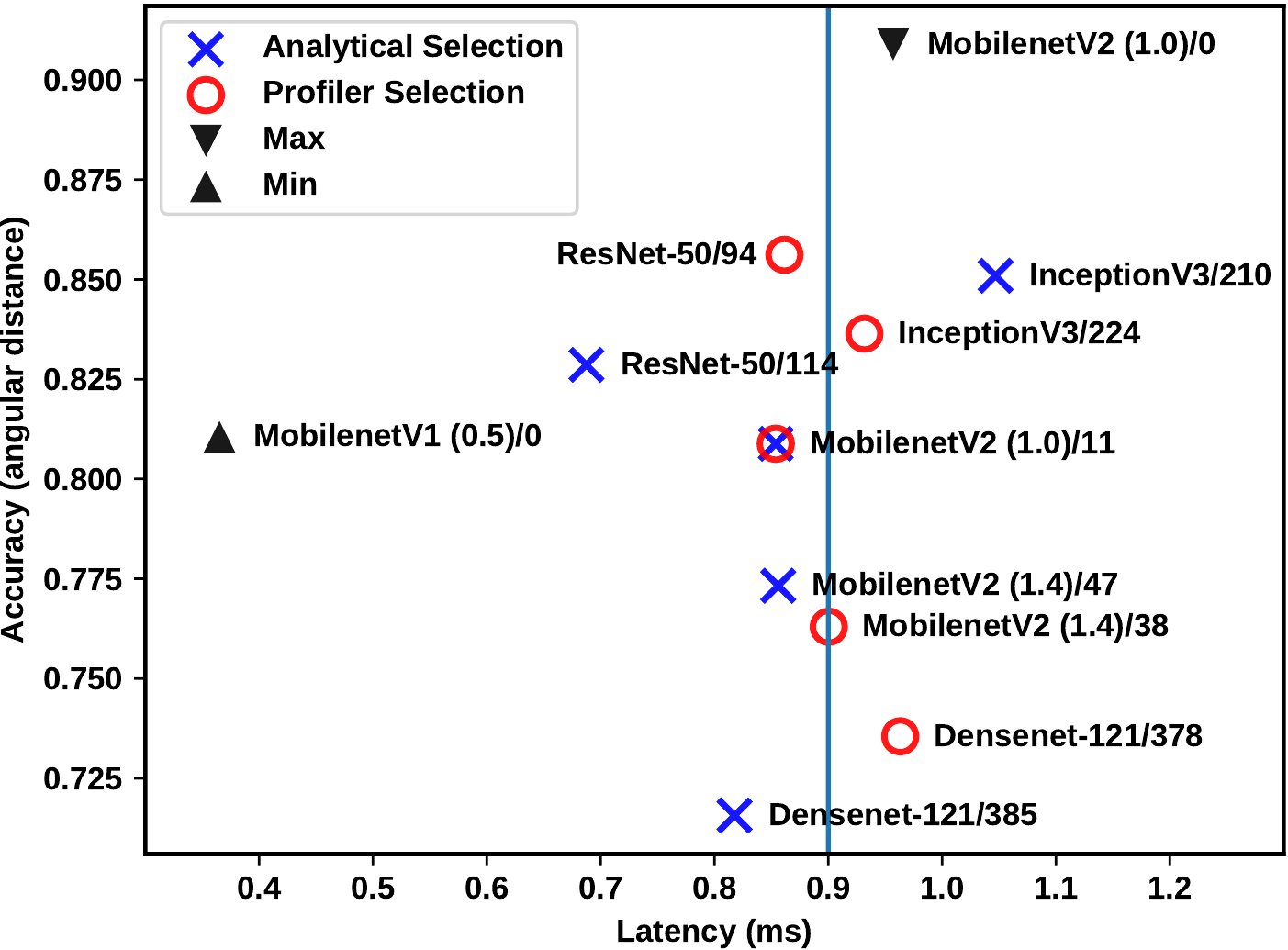}
    \caption{\small Final Selected Networks.}
    \label{fig:final_models}
  \end{minipage}
\end{figure*}

The proposed methodology, NetCut, has been applied to the same networks used in \autoref{sec:layer_removal} to demonstrate its effectiveness at predicting the inference delay and finding more accurate networks based on layer removal. 

After profiling each network, and training the analytical model, we compared the actual measured latency with the predictions. \autoref{fig:res_est} shows the latency estimations for TRNs of ResNet. As seen here, the analytical model is capable of adapting to non-linearities of the actual measured ground truth which can improve the overall accuracy of the prediction. The accuracy of the prediction methods for all networks has been depicted in \autoref{fig:est_erros}. In 2 cases, the analytical model outperforms the profiler-based estimation: ResNet-50 and DenseNet-121. Using the profiler-based approach, the average relative error on all networks was the trivial value of 3.5\% or 0.024 ms. In addition, the error using the analytical model was also negligible. The average relative error was 4.28\% (0.029 ms). To further demonstrate the effectiveness of the RBF kernel, we have examined using linear regression for the same experiments which yielded unacceptable error of 23.81\% (0.092 ms). 

Finally, the aforementioned latency estimation models are employed by NetCut to select TRNs meeting the deadline. The networks selected by profiler- and analytical- based estimation can be found in \autoref{fig:final_models}. As depicted, ResNet/114 and ResNet/94 can be provided as the final network. ResNet/94, can improve accuracy by 5.7\%, and ResNet/114 by 2.2\%, while meeting the prosthetic hand's deadline. This has been achieved while only training 9 additional networks comparing to 148 blockwise candidates. The total time to train networks using blockwise exploration was 183 hours (7 and a half days) on an NVIDIA Tesla K20m, while using NetCut only require 6.7 hours, resulting in a 27x speedup in exploration time.

\section{Conclusion}
In summary, based on the observation that often last layers are problem-specific, we studied the effects of layer removal, and proposed a methodology to explore different TRNs based on that to exploit slack time and increase accuracy. We used a custom dataset designed for recovering the lost ability of amputees in grasping objects executed on a portable embedded system with limited compute power and a tight deadline. We demonstrated that the proposed latency estimations have a trivial error, and based on that, NetCut only trains networks with inference latencies that fall within the deadline. We showed that layer removal can construct TRNs that can have relative improvements of up to $10.43\%$, which combined with NetCut can improve exploration time by 27x.

\section*{Acknowledgment}
We would like to thank Mohammad Nabizadehmashhadtoroghi for design support. This work is partially supported by NSF (CPS-1544895, CPS-1544636, CPS-1544815).

\bibliographystyle{ieeetr}

\bibliography{sample-base}

\begin{thebibliography}{10}

\bibitem{deeplearning}
Y.~LeCun, Y.~Bengio, and G.~Hinton, ``Deep learning,'' {\em Nature}, vol.~521,
  pp.~436--44, 05 2015.

\bibitem{speech}
L.~Deng, J.~Li, J.-T. Huang, K.~Yao, D.~Yu, F.~Seide, M.~Seltzer, G.~Zweig,
  X.~He, J.~Williams, Y.~Gong, and A.~Acero, ``Recent advances in deep learning
  for speech research at microsoft,'' IEEE International Conference on
  Acoustics, Speech, and Signal Processing (ICASSP), May 2013.

\bibitem{alexnet}
A.~Krizhevsky, I.~Sutskever, and G.~E. Hinton, ``Imagenet classification with
  deep convolutional neural networks,'' in {\em Advances in Neural Information
  Processing Systems} (F.~Pereira, C.~J.~C. Burges, L.~Bottou, and K.~Q.
  Weinberger, eds.), vol.~25, pp.~1097--1105, 2012.

\bibitem{medical}
A.~Esteva, B.~Kuprel, R.~Novoa, J.~Ko, S.~Swetter, H.~Blau, and S.~Thrun,
  ``Dermatologist-level classification of skin cancer with deep neural
  networks,'' {\em Nature}, vol.~542, 01 2017.

\bibitem{auto}
C.~Chen, A.~Seff, A.~Kornhauser, and J.~Xiao, ``Deepdriving: Learning
  affordance for direct perception in autonomous driving,'' pp.~2722--2730, 12
  2015.

\bibitem{vivian}
V.~{Sze}, Y.~{Chen}, T.~{Yang}, and J.~S. {Emer}, ``Efficient processing of
  deep neural networks: A tutorial and survey,'' {\em Proceedings of the IEEE},
  vol.~105, no.~12, pp.~2295--2329, 2017.

\bibitem{vgg}
K.~Simonyan and A.~Zisserman, ``Very deep convolutional networks for
  large-scale image recognition,'' in {\em International Conference on Learning
  Representations}, 2015.

\bibitem{resnet50}
K.~He, X.~Zhang, S.~Ren, and J.~Sun, ``Deep residual learning for image
  recognition,'' in {\em Proceedings of the IEEE conference on computer vision
  and pattern recognition}, pp.~770--778, 2016.

\bibitem{densenet}
G.~Huang, Z.~Liu, L.~van~der Maaten, and K.~Q. Weinberger, ``Densely connected
  convolutional networks,'' in {\em Proceedings of the IEEE Conference on
  Computer Vision and Pattern Recognition (CVPR)}, July 2017.

\bibitem{howtransferable}
J.~Yosinski, J.~Clune, Y.~Bengio, and H.~Lipson, ``How transferable are
  features in deep neural networks?,'' in {\em Advances in Neural Information
  Processing Systems 27}, pp.~3320--3328, Curran Associates, Inc., 2014.

\bibitem{QNN}
I.~Hubara, M.~Courbariaux, D.~Soudry, R.~El-Yaniv, and Y.~Bengio, ``Quantized
  neural networks: Training neural networks with low precision weights and
  activations,'' {\em J. Mach. Learn. Res.}, vol.~18, pp.~187:1--187:30, 2017.

\bibitem{struct_pruning}
S.~Anwar, K.~Hwang, and W.~Sung, ``Structured pruning of deep convolutional
  neural networks,'' {\em ACM Journal on Emerging Technologies in Computing
  Systems}, vol.~13, no.~3, p.~1–18, 2017.

\bibitem{mobilenet}
A.~G. Howard, M.~Zhu, B.~Chen, D.~Kalenichenko, W.~Wang, T.~Weyand,
  M.~Andreetto, and H.~Adam, ``Mobilenets: Efficient convolutional neural
  networks for mobile vision applications,'' {\em ArXiv}, vol.~abs/1704.04861,
  2017.

\bibitem{BranchyNet}
S.~Teerapittayanon, B.~McDanel, and H.~Kung, ``Branchynet: Fast inference via
  early exiting from deep neural networks,'' pp.~2464--2469, 12 2016.

\bibitem{edgent}
E.~Li, Z.~Zhou, and X.~Chen, ``Edge intelligence: On-demand deep learning model
  co-inference with device-edge synergy,'' pp.~31--36, 08 2018.

\bibitem{NetAdapt}
T.-J. Yang, A.~Howard, B.~Chen, X.~Zhang, A.~Go, M.~Sandler, V.~Sze, and
  H.~Adam, ``Netadapt: Platform-aware neural network adaptation for mobile
  applications,'' in {\em Computer Vision -- ECCV 2018}, (Cham), pp.~289--304,
  Springer International Publishing, 2018.

\bibitem{cyphy19}
M.~Zandigohar, M.~Han, D.~Erdo{\u{g}}mu{\c{s}}, and G.~Schirner, ``Towards
  creating a deployable grasp type probability estimator for a prosthetic
  hand,'' in {\em Cyber Physical Systems. Model-Based Design} (R.~Chamberlain,
  M.~Edin~Grimheden, and W.~Taha, eds.), (Cham), pp.~44--58, Springer
  International Publishing, 2020.

\bibitem{inceptionV3}
C.~Szegedy, V.~Vanhoucke, S.~Ioffe, J.~Shlens, and Z.~Wojna, ``Rethinking the
  inception architecture for computer vision,'' in {\em Proceedings of the IEEE
  conference on computer vision and pattern recognition}, pp.~2818--2826, 2016.

\bibitem{han2020hands}
M.~Han, S.~Y. G{\"u}nay, G.~Schirner, T.~Pad{\i}r, and D.~Erdo{\u{g}}mu{\c{s}},
  ``Hands: a multimodal dataset for modeling toward human grasp intent
  inference in prosthetic hands,'' {\em Intelligent Service Robotics}, vol.~13,
  no.~1, pp.~179--185, 2020.

\bibitem{quantization}
R.~Krishnamoorthi, ``Quantizing deep convolutional networks for efficient
  inference: A whitepaper,'' {\em ArXiv}, vol.~abs/1806.08342, 2018.

\bibitem{decaf}
J.~Donahue, Y.~Jia, O.~Vinyals, J.~Hoffman, N.~Zhang, E.~Tzeng, and T.~Darrell,
  ``Decaf: A deep convolutional activation feature for generic visual
  recognition,'' in {\em Proceedings of the 31st International Conference on
  International Conference on Machine Learning - Volume 32}, ICML’14,
  p.~I–647–I–655, JMLR.org, 2014.

\end{thebibliography}

\end{document}